\documentclass{article}

\usepackage{microtype}
\usepackage{graphicx}
\usepackage{subcaption}
\usepackage{booktabs} 
\usepackage{caption}
\usepackage[dvipsnames]{xcolor}

\usepackage{hyperref}

\usepackage{amsmath}
\usepackage{bm}

\newcommand\cut[1]{}

\newcommand{\be}{\begin{equation}}
\newcommand{\ee}{\end{equation}}
\newcommand{\bea}{\begin{eqnarray}}
\newcommand{\eea}{\end{eqnarray}}
\newcommand{\beaa}{\begin{eqnarray*}}
\newcommand{\eeaa}{\end{eqnarray*}}

\DeclareMathAlphabet{\mathpzc}{OT1}{pzc}{m}{n}

\newcommand{\modelname}{Constellation}
\DeclareMathOperator{\im}{\mathbf{x}} 
\DeclareMathOperator{\obj}{\mathbf{o}} 
\DeclareMathOperator{\abs}{\mathbf{a}} 
\DeclareMathOperator{\rel}{\mathbf{r}} 
\DeclareMathOperator{\lan}{\mathbf{l}} 

\DeclareMathOperator{\absh}{\mathbf{\hat{a}}} 

\DeclareMathOperator{\rels}{\mathbf{\tilde{r}}} 

\DeclareMathOperator{\muq}{\bm{\mu_q}} 
\DeclareMathOperator{\sigq}{\bm{\sigma_q}} 
\DeclareMathOperator{\mus}{\bm{\mu_s}}  
\DeclareMathOperator{\sigs}{\bm{\sigma_s}}  

\DeclareMathOperator{\mask}{\mathbf{m}} 

\DeclareMathAlphabet{\mathpzc}{OT1}{pzc}{m}{n}

\usepackage[accepted]{icml2021}

\icmltitlerunning{Constellation: Learning relational abstractions over objects for compositional imagination}

\begin{document}

\twocolumn[
    \icmltitle{Constellation: Learning relational abstractions over objects\\ for compositional imagination}
    
    
    
    \icmlsetsymbol{equal}{*}
    
    \begin{icmlauthorlist}
    \icmlauthor{James C.R. Whittington}{ox,exdm}
    \icmlauthor{Rishabh Kabra}{dm}
    \icmlauthor{Loic Matthey}{dm}
    \icmlauthor{Christopher P. Burgess}{wz,exdm}
    \icmlauthor{Alexander Lerchner}{dm}
    
    
    \end{icmlauthorlist}
    
    \icmlaffiliation{ox}{University of Oxford}
    \icmlaffiliation{dm}{DeepMind}
    \icmlaffiliation{wz}{Wayve}
    \icmlaffiliation{exdm}{Work done at DeepMind}
    
    \icmlcorrespondingauthor{JCRW}{jcrwhittington@gmail.com}
    \icmlcorrespondingauthor{AL}{lerchner@google.com}
    
    \icmlkeywords{Machine Learning, ICML}
    
    \vskip 0.3in
]



\printAffiliationsAndNotice{}  

\begin{abstract}
Learning structured representations of visual scenes is currently a major bottleneck to bridging perception with reasoning. While there has been exciting progress with slot-based models, which learn to segment scenes into sets of objects, learning configurational properties of entire groups of objects is still under-explored. To address this problem, we introduce \emph{\modelname{}}, a network that learns relational abstractions of static visual scenes, and generalises these abstractions over sensory particularities, thus offering a potential basis for abstract relational reasoning. We further show that this basis, along with language association, provides a means to imagine sensory content in new ways. This work is a first step in the explicit representation of visual relationships and using them for complex cognitive procedures.
\end{abstract}

\section{Introduction}
\label{sec:intro}

\begin{figure}[t]
    \centering
    \includegraphics[width=0.7\columnwidth]{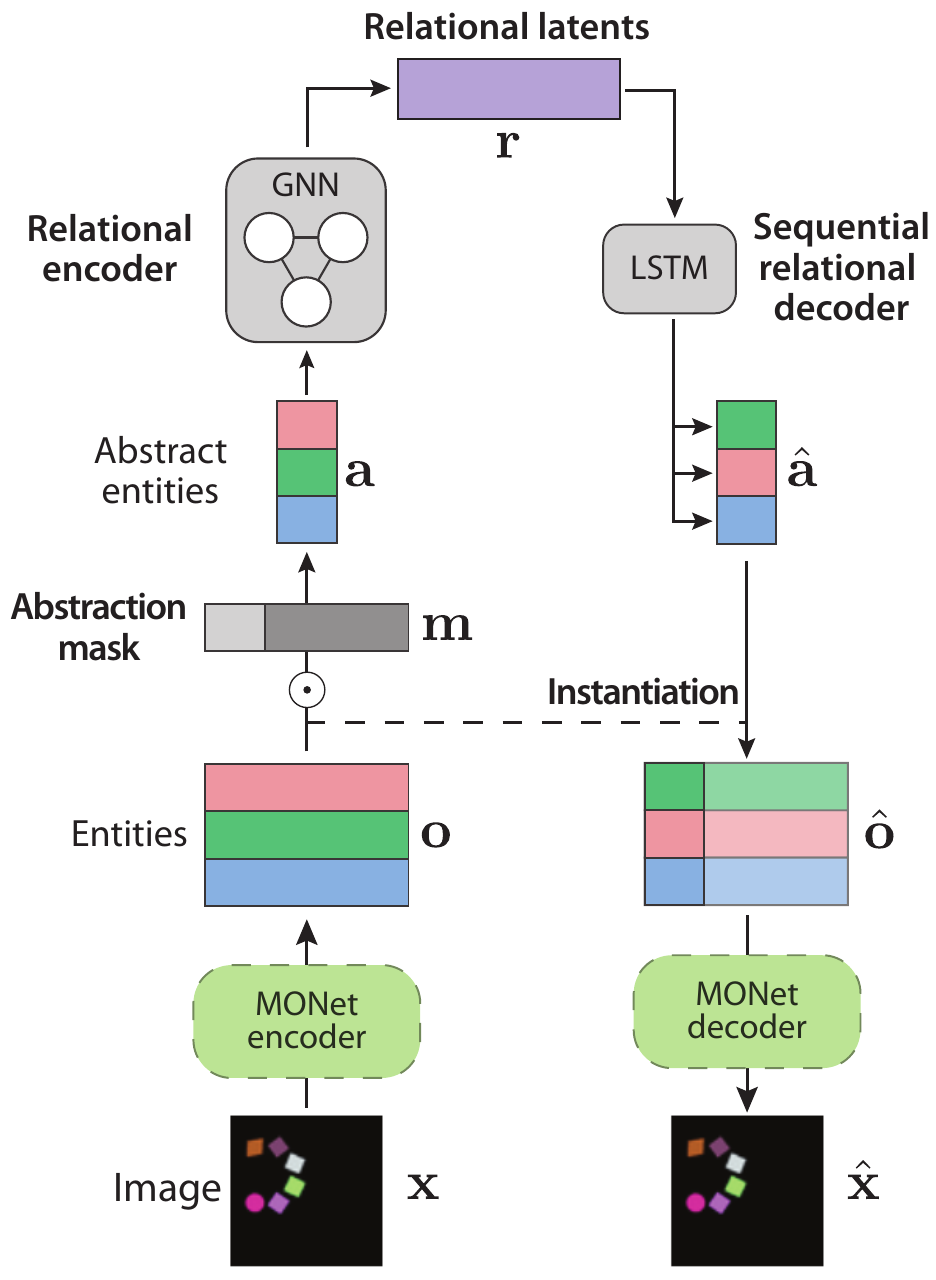}
    \caption{\textbf{Constellation model overview}. A \textit{pre-trained} MONet encoder is used to decompose scenes of objects into slots with a common feature representation. Each slot either contains an ``object'' representation or an ``empty'' representation. Features irrelevant to relationships are abstracted away by applying a learned mask, \( \mask \). Relational structure, \( \rel \), is inferred from abstracted objects, \( \abs \), via a permutation invariant encoder. Finally, \modelname{} decodes the abstracted objects using an LSTM. Abstracted away features can be 'filled in' using encoded objects \(\obj\) and passed through the MONet decoder for visual analysis.}
    \label{fig:model}
    \vspace{-15pt}
\end{figure}

Humans and other animals understand the world at an abstracted level. From a line of ducks or a queue of people, we understand the common abstraction of ``lineness''. This abstract lineness, a concept divorced from sensory particulars (ducks or humans), is what we reason over when saying ``that line is long'' or ``Alice is two ahead of me in the queue''. Aside from offering a basis to reason over, knowledge of abstract concepts allows understanding and imagination of never-seen-before sensory configurations - you can understand and imagine a house built from monkeys, because you separately understand the two components; monkeys and how houses are configured. Recent approaches to learning abstractions suggest different parts of sensory experience be represented separately i.e. in a factorised fashion \cite{greff2020, Higgins2017, Higgins2018a}. Factorised sensory representations are easily re-combined to represent novel experiences, and thus afford out-of-distribution generalisation and transfer \cite{dittadi2020, steenkiste2019, Higgins2017}. As in the line example, sensory scenes often have underlying relational structure. Factorising structure from sensory content, similarly provides a means to generalise relational knowledge \cite{Behrens2018, Whittington2018, Whittington2020}.

Using such \textit{relational abstraction}, we aim to learn and understand concepts of relationships between objects, and show this understanding affords imagination of objects in never-seen-before configurations. For relational abstraction to work, sensory scenes must contain relational regularities i.e. there exists subsets of object properties (e.g. x, y location) described by generative relational factors. For lines of ducks (or arbitrary objects), these underlying factors describing lines might consist of number of objects, length of line, curviness etc. The problem of relational abstraction can thus be approached as an inverse problem; inferring these generative relational factors from data. We demonstrate this approach of relational abstraction is sufficient to build \emph{\modelname{}}; a network that learns and understands relational concepts. We then show that ``language'' symbols (e.g. 'circle') can be bound to abstract relational concepts, which allows for compositional (re-)imagination of never-seen-before object configurations (e.g. a line of hats (re-)imagined in a circle).

\begin{figure}[!t]
    \centering
    \includegraphics[width=0.95\linewidth]{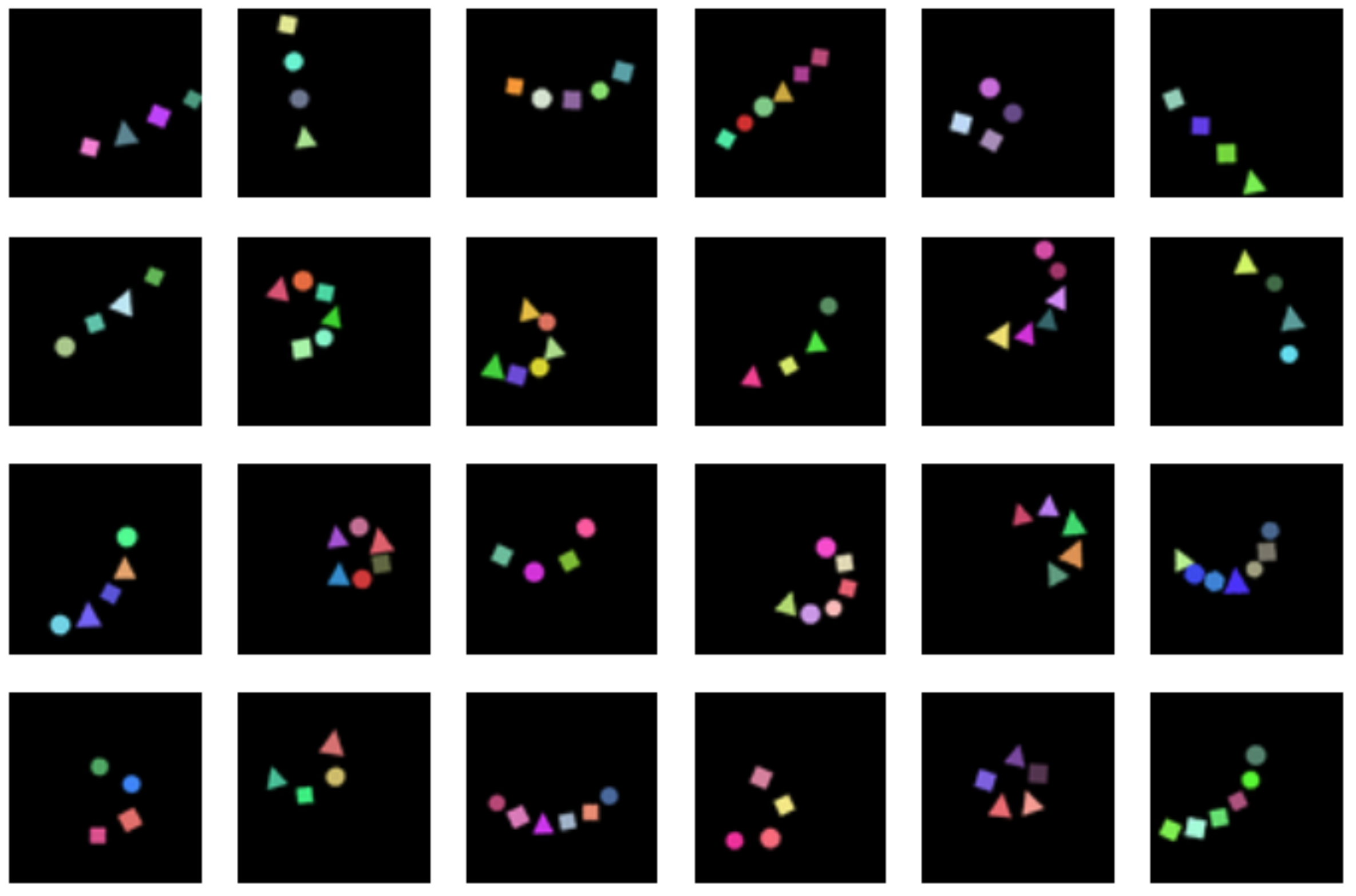}
    \caption{\textbf{Dataset with relational structure between objects.} Dataset of images containing multiple objects in ``super-structures''. Each ``super-structure'' is defined by a set of generative factors: position, number of objects, orientation, curviness. The visual features of each object are randomly chosen.} 
    \label{fig:dataset}
    \vspace{-15pt}
\end{figure}

\section{\modelname{} - a network for relational abstraction and imagination}\label{S:model}

To learn about relational knowledge and infer relational generative factors, we use generative modelling, and in particular, the variational autoencoder framework (VAEs; \cite{Kingma2013a, Rezende2014}). A key point in our learning of abstractions is that we do not try to predict all factors of the objects (see dataset examples in Figure \ref{fig:dataset}). Instead we only generate a subset of each object's properties: i.e. all the x, y positions of the objects and nothing else. This subset, an abstraction over all object features sufficient to capture 2D arrangements, is learned by our model in an unsupervised manner. Abstracting over object features, and only reconstructing these features, allows the network to learn the generative \textit{relational} factors \textit{alone}. This induces an implicit factorisation; relational information of the scene is separated from the sensory content. It is this separation that supports generalisation and imagination. This separation, however, provides a quandary - pure relational knowledge lacks sensory knowledge, so how can we imagine a full visual scene? To address this, we use a ``filling in'' procedure that takes the partially-specified generated objects, compares them to the full objects in the current scene and fills in the remaining properties accordingly (details discussed later). 

To obtain object representations in the first place, we use a \textit{scene decomposition} network, in particular, MONet \cite{Burgess2019a}). MONet takes as input a visual scene (\( \im \)), containing variable numbers of objects, and encodes the objects into entities (\( \obj_i \)) with a common \textit{disentangled} representational format (Figure~\ref{fig:model} bottom). We will consistently use subscript \(i\) to index different object vectors, and subscript \( j \) to index an object feature representation. Slots can either consist of an object representation or an ``empty'' representation.  To abstract over object features (Figure~\ref{fig:model} middle), we use a learned mask, \( \mask \), where \( \mask \) must learn to ignore irrelevant features (colour, size etc), and keep the relevant ones (x,y). The mask is applied to each object (slot) element-wise, to obtain an \textit{abstract} entity \( \abs_i = \obj_i  \odot \mask \), where \( m_j > 0 \) and \( \sum_j m_j = 1 \). 

\textbf{Generative model.}
While the procedure from relational generative factors, \( \rel \), to visual scene (\( \im \)) can be framed as a hierarchical generative model, here we use a pre-trained MONet module to provide a slotted scene representation, \( \obj_i \). Thus, in this work, we need only reconstruct \( \abs\), rather than \( \im \), i.e. we generate only a subset of each object's feature representations (subset \( \abs_i \) of objects \( \obj_i \)) and thus want to maximise the log-probability, \( \ln P(\abs) =  \sum_i \ln P ( \abs_i ) \). We assume a generative model \( P (\abs, \rel ) =  P( \abs \mid \rel) P (\rel) \), where \( \rel \) is the relational generative factors. \( P(\rel) \) is Gaussian, and \(  P(\abs_i \mid \rel) \sim \mathcal{N} (\bm{\mu}_i, \bm{\sigma}_i ) \). The mean vectors, \( \bm{\mu}_i \), for each object (slot) are produced sequentially via an LSTM \cite{Hochreiter1997}; \( \bm{\mu}_i =  f( LSTM_i ( g( \rel ), LSTM_{i-1} ) ) \), where \( f( \cdots ) \) and \( g ( \cdots) \) are neural networks (Figure~\ref{fig:model} Right).

\textbf{Inference network.}
For the recognition model for relational structure, we choose a diagonal Gaussian latent posterior \( Q ( \rel \mid \abs ) \sim \mathcal{N} (\muq, \sigq ) \). Since the slots containing \( \abs_i \) may be randomly ordered, we use a permutation invariant encoder (e.g. a graph neural network; GNN; \cite{Battaglia2018}). We use the ``globals'' of the GNN to parameterise \( \muq \) and \( \sigq \) of the variational posterior.

\textbf{Filling in abstracted visual features.}
Generating abstracted object features is simple in ancestral sampling (from \( \rel \)) as they can sampled from MONet's prior. However, when encoding an image we need to fill back in the correct object features. To do this, we use a non-parametric matching algorithm (Hungarian algorithm; \cite{Kuhn1955}). After pairing up encoded object \( \obj_i \) to decoded ``abstract object'' \( \absh_i \), the masked out features are replaced using \( \obj_i \) (Figure \ref{fig:model}, ``instantiation'' arrows). 

\textbf{Training.} We use an augmented VAE loss. In particular: 

\textit{Permutation invariant reconstruction error}: 
Our main goal is to accurately reconstruct abstracted relational features \( \absh_i \) (as we use a pre-trained MONet, we do not reconstruct \(\im\) directly).
However, the encoded and generated objects, \( \abs_i \) and \( \absh_i \), may be ordered differently, preventing simple slot pairing. Instead we use the Hungarian matching algorithm \cite{Kuhn1955} to pair slots for the reconstruction error.
\( L_{rec} = \frac{1}{2} \sum_{(i,j)-pairs} \left| \right|  \abs_i - \absh_j  \left| \right| ^2 \).

\textit{Disentangling pressure}: 
To encourage disentangling (to facilitate subsequent language association and compositional imagination), we use a \(\beta\)-VAE \cite{Higgins2017} KL regularizer:
\( L_{reg} = \beta D_{\mathrm{KL}} \left(  Q(\rel \mid \im) \left| \right| P(\rel) \right) \).

\textit{Mask entropy}:
We use a mask entropy loss to prevent the abstraction mask collapsing onto a trivial single feature dimension;
\( L_{entropy} = - \sum_j m_j \ln m_j \).

\textit{Conditioning term}: Since gradients cannot flow through masked-out features, it is difficult to ``unlearn'' an abstraction mask. This is problematic if the incorrect features are abstracted initially. To remedy this, and allow gradient information to flow through masked object features, we introduce an additional loss
\( L_{condition} = \sum_j (1 - m_j)~|~L^{*}_{rec_j} - \gamma_j~| \).
\( \gamma_j \) is learnable and predicts the reconstruction loss for that \textit{feature}; \(L^{*}_{rec_j}\). \(L^{*}_{rec_j}\) is obtained using the same pairings as \(L_{rec_j}\), but using unmasked representation so that \( \absh \) and \( \obj \) have the same scaling i.e. \( L^{*}_{rec_j} = \frac{1}{2} \sum_{(i,j)-pairs} \left( \obj_i - \frac{\absh_j}{m_j} \right) ^2 \). \( L_{condition} \) is decayed to 0 throughout training.

\textit{Re-ordering loss}:
Parsimonious structured representations would generate nearby objects in sequence i.e. for a line, starting at one end and finishing at the other. To encourage this, we add a penalty to the difference between successively generated objects.
\( L_{reorder} = \sum_{i=1} \left| \right| \absh_i  - \absh_{i-1}  \left| \right| ^2 \).

We optimise \( L = L_{reg} + L_{rec} + L_{entropy} + L_{re-order} + L_{condition} \) with Adam \cite{Kingma2014b} and GECO \cite{Rezende2018} - a constrained optimisation method. We use a learning rate of \(1e-3\) annealed to \(1e-4\). 

\subsection{Learning disentangled relational knowledge}

\begin{figure}[!t]
    \centering
    \includegraphics[width=\linewidth]{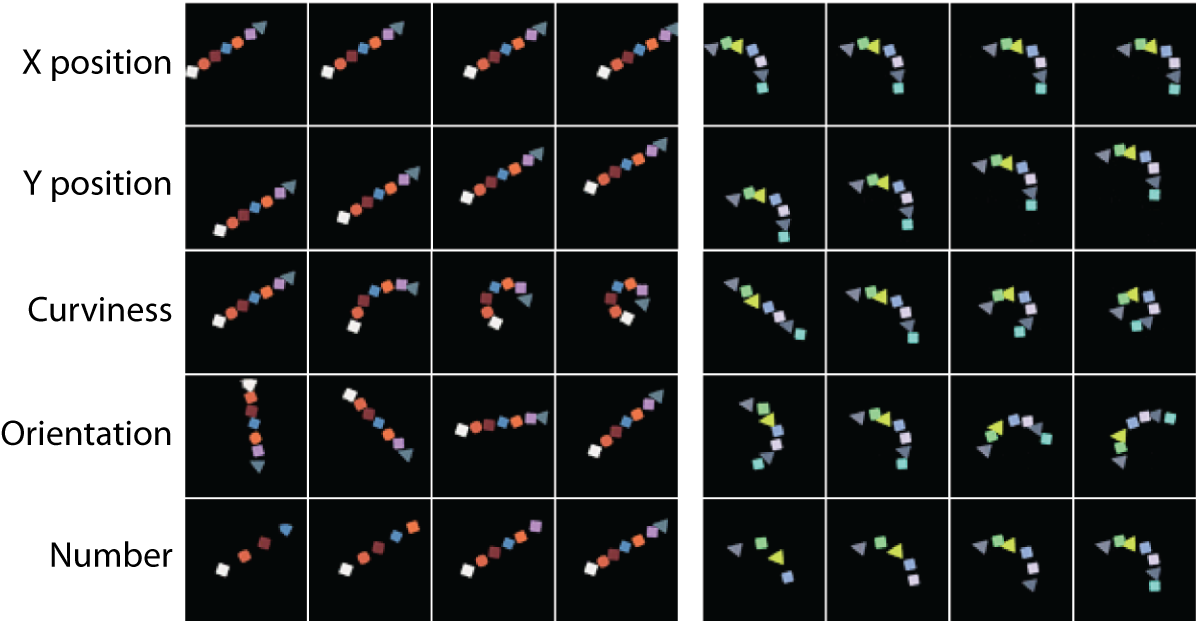}
    \caption{\textbf{\modelname{} learns disentangled relational knowledge that generalises over sensory properties.} Latent traversals of two different input images, 4 traversals points are shown for each latent dimension. Each latent dimension captures a single relational factor, that smoothly changes across the latent sweep. These relational factors are the same no matter the sensory particulars of the objects i.e. they generalise. All other relational latents are unused and have no effects when traversed.}
    \label{fig:traversals}
\end{figure}

\begin{figure}[!t]
    \centering
    \includegraphics[width=0.9\linewidth]{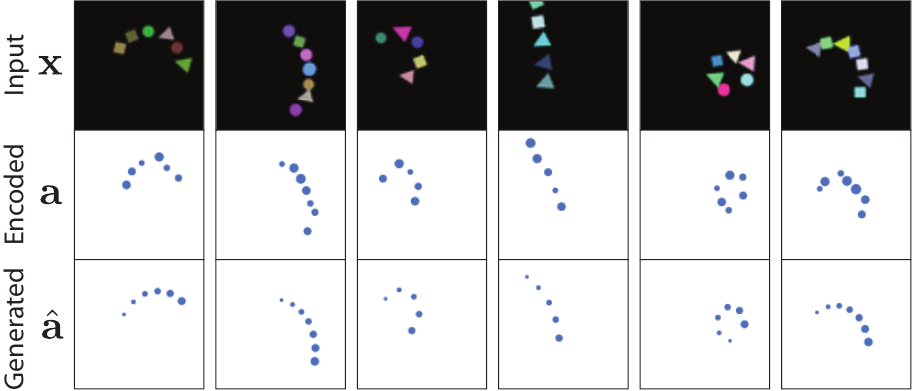}
    \caption{\textbf{Objects re-ordered according to prior learned relational knowledge.} \textbf{Top:} Input image. \textbf{Middle:} MONet latent dimensions corresponding to x,y. Size of point corresponds to order in the sequence. \textbf{Bottom:} \modelname{} decoded x,y dimensions, where objects are re-ordered according to the learned line structure (see size of points compared to the panels above).} 
    \label{fig:order}
    \vspace{-5pt}
\end{figure}

We visualise the latent traversals of \modelname{} by encoding an image and sweeping over single relational latent dimensions (Figure \ref{fig:traversals}). \modelname{} learns the five relational factors in separate latents, successfully disentangling them and leaving the rest of its features unused (see Appendix for a MIG analysis).
\modelname{} learns parsimonious relational knowledge representations that generate objects in an order obeying the structural properties of the line (Figure \ref{fig:order}). Note the MONet representations are unordered. This re-ordering is encouraged with \( L_{reorder} \), though it is observed (to a lesser extent) without this loss. Finally, we note the relational factors generalise over different sensory objects; relational knowledge is factorised from sensory particulars.

\begin{figure*}[!t]
    \centering
    \includegraphics[width=\linewidth]{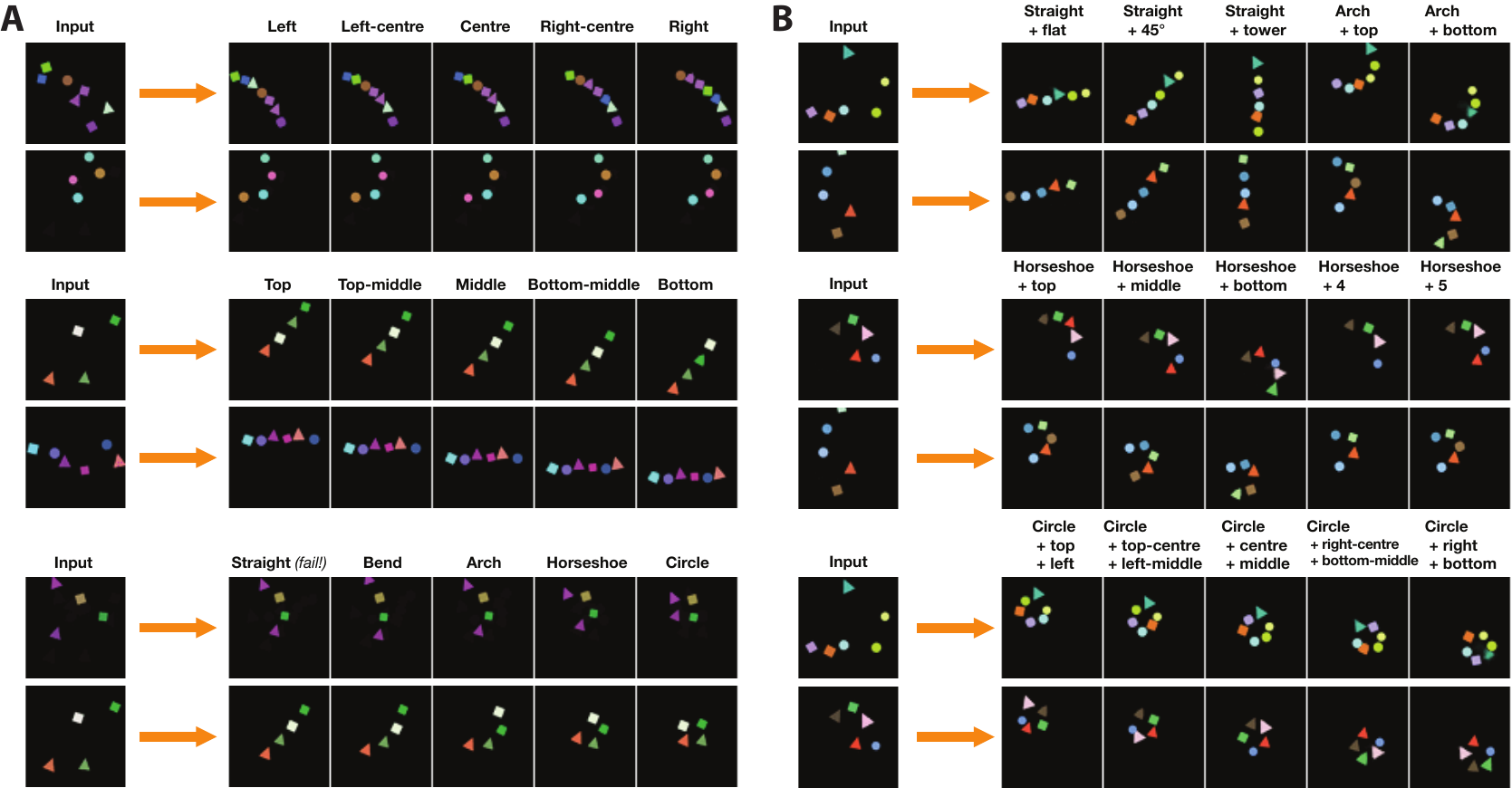}
    \caption{\textbf{Imagining novel objects in different configurations.} \textbf{A} After encoding the relationships of a visual scene, new structural constraints can be imposed (above each image). \textbf{B} Similarly, several constraints can be combined and enforced.}
    \label{fig:imaginations}
\end{figure*}

\section{Imagination from language}\label{S:scan}

Imagination is crucial to intelligent behaviour. When instructed to build a tower out of blocks scattered on the floor, first imagining how those blocks would look like in a tower is useful as it acts as a conceptual goal state. This, however, relies on knowing what 'tower' means - i.e. one needs to understand the concept of a vertical line and that it is labelled a 'tower'. Labelling structural forms is necessary for systems to be instructed to imagine X out of Y e.g. a house out of monkeys. We adapt SCAN \cite{Higgins2018}, a network architecture that binds language symbols, \( \lan \), to abstractions over disentangled representations, \( \rel \) (see Appendix for details). 

\textbf{Constraining relational knowledge with language.}
This association allows ``ancestral'' generation of visual scenes based on language input \cite{Higgins2018}. More interestingly though, we can re-interpret scenes of objects under different relational constraints e.g. from a relational encoding of 5 squares in a straight line, we can impose 'circle' from language and re-imagine those 5 squares in a circle. To impose structural constraints on relational embeddings, we encode \( \abs \) to get \( \rel \), sample the predicted \( \rels \) from \( \lan \), then overwrite \( \rel \) with the elements of \( \rels \) that have a high relevance for that relational feature (i.e. low \(\sigs\); see Appendix for details).
The elements with low \(\sigs\) are the relational factors described by language, thus this provides a mechanism for imposing ``imagined'' structure on existing inferred structure (i.e. turns a straight line of objects into a circle or objects). Examples of imagination are shown in Figure~\ref{fig:imaginations}A. Visual scenes of randomly placed objects are re-imagined in lines of various forms.

\textbf{Compositional constraints.}
Our model can perform "compositional imagination" by providing language input corresponding to combinations of properties. For example, 'left circle' should re-imagine the objects into a circle on the left. Figure~\ref{fig:imaginations}B shows this for a variety of different language symbols, and for compositions of 2 or 3 language symbols.

\section{Related work}\label{S:related}

Using permutation invariant encoders within the VAE framework for relational understanding of scenes has been explored with capsule networks \cite{Kosiorek2019}. There the impetus is on exploiting view-point invariances of intrinsic geometric relationships of objects. Here we are instead interested in explicitly representing the relational structure of a scene of objects and using this representation to enable complex cognitive operations like imagination. Performing inference on novel objects as recombination of sub-objects through a generative model has also been explored in \cite{Lake2015}, and in \cite{Kemp2008} the inference over particular structural forms are considered.

\section{Conclusion}\label{S:conclusion}

We introduced \modelname{}, a network that learns structural regularities among sets of objects and represents these relationships, in a disentangled manner, divorced from sensory particularities. We have shown how these latent embeddings can be associated to language, which allows re-imagination of the current set of objects in new ways.

While we believe this work provides an exciting step in learning relational abstractions and enacting imaginations, there is still much work to be done. For example, we use synthetically generated datasets that lack the visual complexity or noise of real life scenes. Nevertheless, our work provides a platform for further research; explicitly representing temporal relationships of stereotyped interacting bodies (e.g. courting or avoidance behaviours), imagination-based conceptual goal states, analogical reasoning on structures, and hierarchical scene decomposition, to name just a few.

\bibliographystyle{icml2021}
\bibliography{references}

\begin{thebibliography}{21}
\providecommand{\natexlab}[1]{#1}
\providecommand{\url}[1]{\texttt{#1}}
\expandafter\ifx\csname urlstyle\endcsname\relax
  \providecommand{\doi}[1]{doi: #1}\else
  \providecommand{\doi}{doi: \begingroup \urlstyle{rm}\Url}\fi

\bibitem[Battaglia et~al.(2018)Battaglia, Hamrick, Bapst, Sanchez-Gonzalez,
  Zambaldi, Malinowski, Tacchetti, Raposo, Santoro, Faulkner, Gulcehre, Song,
  Ballard, Gilmer, Dahl, Vaswani, Allen, Nash, Langston, Dyer, Heess, Wierstra,
  Kohli, Botvinick, Vinyals, Li, and Pascanu]{Battaglia2018}
Battaglia, P.~W., Hamrick, J.~B., Bapst, V., Sanchez-Gonzalez, A., Zambaldi,
  V., Malinowski, M., Tacchetti, A., Raposo, D., Santoro, A., Faulkner, R.,
  Gulcehre, C., Song, F., Ballard, A., Gilmer, J., Dahl, G., Vaswani, A.,
  Allen, K., Nash, C., Langston, V., Dyer, C., Heess, N., Wierstra, D., Kohli,
  P., Botvinick, M., Vinyals, O., Li, Y., and Pascanu, R.
\newblock {Relational inductive biases, deep learning, and graph networks}.
\newblock \emph{arXiv}, pp.\  1--38, 2018.
\newblock URL \url{http://arxiv.org/abs/1806.01261}.

\bibitem[Behrens et~al.(2018)Behrens, Muller, Whittington, Mark, Baram,
  Stachenfeld, and Kurth-nelson]{Behrens2018}
Behrens, T. E.~J., Muller, T.~H., Whittington, J. C.~R., Mark, S., Baram,
  A.~B., Stachenfeld, K.~L., and Kurth-nelson, Z.
\newblock {What Is a Cognitive Map? Organizing Knowledge for Flexible
  Behavior}.
\newblock \emph{Neuron}, 100\penalty0 (2):\penalty0 490--509, 2018.
\newblock ISSN 0896-6273.
\newblock \doi{10.1016/j.neuron.2018.10.002}.
\newblock URL \url{https://www.cell.com/neuron/fulltext/S0896-6273(18)30856-0}.

\bibitem[Burgess et~al.(2019)Burgess, Matthey, Watters, Kabra, Higgins,
  Botvinick, and Lerchner]{Burgess2019a}
Burgess, C.~P., Matthey, L., Watters, N., Kabra, R., Higgins, I., Botvinick,
  M., and Lerchner, A.
\newblock {MONet: Unsupervised Scene Decomposition and Representation}.
\newblock \emph{arXiv}, pp.\  1--22, 2019.
\newblock URL \url{http://arxiv.org/abs/1901.11390}.

\bibitem[Dittadi et~al.(2020)Dittadi, Tr{\"{a}}uble, Locatello, W{\"{u}}thrich,
  Agrawal, Winther, Bauer, and Sch{\"{o}}lkopf]{dittadi2020}
Dittadi, A., Tr{\"{a}}uble, F., Locatello, F., W{\"{u}}thrich, M., Agrawal, V.,
  Winther, O., Bauer, S., and Sch{\"{o}}lkopf, B.
\newblock On the transfer of disentangled representations in realistic
  settings.
\newblock \emph{CoRR}, abs/2010.14407, 2020.
\newblock URL \url{https://arxiv.org/abs/2010.14407}.

\bibitem[Greff et~al.(2020)Greff, van Steenkiste, and Schmidhuber]{greff2020}
Greff, K., van Steenkiste, S., and Schmidhuber, J.
\newblock On the binding problem in artificial neural networks.
\newblock \emph{CoRR}, abs/2012.05208, 2020.
\newblock URL \url{https://arxiv.org/abs/2012.05208}.

\bibitem[Higgins et~al.(2017)Higgins, Matthey, Pal, Burgess, Glorot, Botvinick,
  Mohamed, and Lerchner]{Higgins2017}
Higgins, I., Matthey, L., Pal, A., Burgess, C., Glorot, X., Botvinick, M.,
  Mohamed, S., and Lerchner, A.
\newblock {{$\beta$}-VAE: Learning basic visual concepts with a constrained
  variational framework}.
\newblock \emph{International Conference on Learning Representations}, 0, 7
  2017.

\bibitem[Higgins et~al.(2018{\natexlab{a}})Higgins, Amos, Pfau, Racaniere,
  Matthey, Rezende, and Lerchner]{Higgins2018a}
Higgins, I., Amos, D., Pfau, D., Racaniere, S., Matthey, L., Rezende, D., and
  Lerchner, A.
\newblock {Towards a Definition of Disentangled Representations}.
\newblock \emph{arXiv}, pp.\  1--29, 12 2018{\natexlab{a}}.
\newblock URL \url{http://arxiv.org/abs/1812.02230}.

\bibitem[Higgins et~al.(2018{\natexlab{b}})Higgins, Sonnerat, Matthey, Pal,
  Burgess, Bosnjak, Shanahan, Botvinick, Hassabis, and Lerchner]{Higgins2018}
Higgins, I., Sonnerat, N., Matthey, L., Pal, A., Burgess, C.~P., Bosnjak, M.,
  Shanahan, M., Botvinick, M., Hassabis, D., and Lerchner, A.
\newblock {SCAN: Learning Hierarchical Compositional Visual Concepts}.
\newblock \emph{arXiv}, pp.\  1--24, 2018{\natexlab{b}}.
\newblock ISSN 1346-9843.
\newblock \doi{10.1186/s12884-017-1520-4}.
\newblock URL \url{http://arxiv.org/abs/1707.03389}.

\bibitem[Hochreiter \& Schmidhuber(1997)Hochreiter and
  Schmidhuber]{Hochreiter1997}
Hochreiter, S. and Schmidhuber, J.
\newblock {Long Short-term Memory}.
\newblock \emph{Neural Computation}, 9\penalty0 (8):\penalty0 17351780, 1997.

\bibitem[Kemp \& Tenenbaum(2008)Kemp and Tenenbaum]{Kemp2008}
Kemp, C. and Tenenbaum, J.~B.
\newblock {The discovery of structural form}.
\newblock \emph{Proceedings of the National Academy of Sciences}, 105\penalty0
  (31):\penalty0 10687--10692, 2008.
\newblock ISSN 0027-8424.
\newblock \doi{10.1073/pnas.0802631105}.
\newblock URL \url{http://www.pnas.org/cgi/doi/10.1073/pnas.0802631105}.

\bibitem[Kingma \& Ba(2014)Kingma and Ba]{Kingma2014b}
Kingma, D.~P. and Ba, J.~L.
\newblock {Adam: A Method for Stochastic Optimization}.
\newblock \emph{arXiv preprint arxiv:1412.6980}, 0, 2014.
\newblock ISSN 09252312.
\newblock \doi{http://doi.acm.org.ezproxy.lib.ucf.edu/10.1145/1830483.1830503}.
\newblock URL \url{http://arxiv.org/abs/1412.6980}.

\bibitem[Kingma \& Welling(2013)Kingma and Welling]{Kingma2013a}
Kingma, D.~P. and Welling, M.
\newblock {Auto-Encoding Variational Bayes}.
\newblock \emph{arXiv preprint arXiv:1312.6114}, 0\penalty0 (Ml):\penalty0
  1--14, 2013.
\newblock ISSN 1312.6114v10.
\newblock \doi{10.1051/0004-6361/201527329}.
\newblock URL \url{http://arxiv.org/abs/1312.6114}.

\bibitem[Kosiorek et~al.(2019)Kosiorek, Sabour, Teh, and Hinton]{Kosiorek2019}
Kosiorek, A.~R., Sabour, S., Teh, Y.~W., and Hinton, G.~E.
\newblock {Stacked Capsule Autoencoders}.
\newblock \emph{arXiv}, 2019.
\newblock URL \url{http://arxiv.org/abs/1906.06818}.

\bibitem[Kuhn(1955)]{Kuhn1955}
Kuhn, H.~W.
\newblock {The Hungarian method for the assignment problem}.
\newblock \emph{Naval Research Logistics Quarterly}, 2\penalty0 (1-2):\penalty0
  83--97, 3 1955.
\newblock ISSN 00281441.
\newblock \doi{10.1002/nav.3800020109}.
\newblock URL \url{http://doi.wiley.com/10.1002/nav.3800020109}.

\bibitem[Lake et~al.(2015)Lake, Salakhutdinov, and Tenenbaum]{Lake2015}
Lake, B.~M., Salakhutdinov, R., and Tenenbaum, J.~B.
\newblock {Human-level concept learning through probabilistic program
  induction}.
\newblock \emph{Science}, 350\penalty0 (6266):\penalty0 1332--1338, 12 2015.
\newblock ISSN 10959203.
\newblock \doi{10.1126/science.aab3050}.

\bibitem[Rezende \& Viola(2018)Rezende and Viola]{Rezende2018}
Rezende, D.~J. and Viola, F.
\newblock {Taming VAEs}.
\newblock \emph{arXiv}, 2018.
\newblock URL \url{http://arxiv.org/abs/1810.00597}.

\bibitem[Rezende et~al.(2014)Rezende, Mohamed, and Wierstra]{Rezende2014}
Rezende, D.~J., Mohamed, S., and Wierstra, D.
\newblock {Stochastic Backpropagation and Approximate Inference in Deep
  Generative Models}.
\newblock \emph{arXiv preprint arXiv:1401.4082}, 0, 1 2014.
\newblock ISSN 10495258.
\newblock \doi{10.1051/0004-6361/201527329}.
\newblock URL \url{http://arxiv.org/abs/1401.4082}.

\bibitem[van Steenkiste et~al.(2019)van Steenkiste, Locatello, Schmidhuber, and
  Bachem]{steenkiste2019}
van Steenkiste, S., Locatello, F., Schmidhuber, J., and Bachem, O.
\newblock Are disentangled representations helpful for abstract visual
  reasoning?
\newblock \emph{CoRR}, abs/1905.12506, 2019.
\newblock URL \url{http://arxiv.org/abs/1905.12506}.

\bibitem[Watters et~al.(2019)Watters, Matthey, Borgeaud, Kabra, and
  Lerchner]{spriteworld19}
Watters, N., Matthey, L., Borgeaud, S., Kabra, R., and Lerchner, A.
\newblock Spriteworld: A flexible, configurable reinforcement learning
  environment.
\newblock https://github.com/deepmind/spriteworld/, 2019.
\newblock URL \url{https://github.com/deepmind/spriteworld/}.

\bibitem[Whittington et~al.(2018)Whittington, Muller, Mark, Barry, and
  Behrens]{Whittington2018}
Whittington, J. C.~R., Muller, T.~H., Mark, S., Barry, C., and Behrens, T.
  E.~J.
\newblock {Generalisation of structural knowledge in the hippocampal-entorhinal
  system}.
\newblock \emph{Advances in Neural Information Processing Systems 31},
  31:\penalty0 8493--8504, 2018.

\bibitem[Whittington et~al.(2020)Whittington, Muller, Mark, Barry, Burgess,
  Behrens, Chen, Barry, Burgess, and Behrens]{Whittington2020}
Whittington, J. C.~R., Muller, T.~H., Mark, S., Barry, C., Burgess, N.,
  Behrens, T. E.~E., Chen, G., Barry, C., Burgess, N., and Behrens, T. E.~E.
\newblock {The Tolman-Eichenbaum Machine: Unifying Space and Relational Memory
  through Generalization in the Hippocampal Formation}.
\newblock \emph{Cell}, 183\penalty0 (5):\penalty0 1249--1263, 11 2020.
\newblock ISSN 00928674.
\newblock \doi{10.1016/j.cell.2020.10.024}.

\end{thebibliography}


\clearpage

\appendix




\section{Dataset}
\label{sec:dataset}

We designed a dataset, using Spriteworld \cite{spriteworld19}, which consists of visual scenes of objects with regular structural properties. In particular, we chose the relational structure to be between object positions, i.e. the arrangements of the objects are all variants of lines, where the line has varying properties: number of objects, curviness of line, position, and orientation. Each object within the `super-structure' is generated from an additional set of factors; colour, shape, size, position, orientation. These factors are randomly sampled, except for the position factors, which are set according to the relational factors detailed above. See Figure \ref{fig:dataset} for example samples of this dataset. This dataset thus conforms to our desiderata - consistent relational structure across different sensory objects. Though we chose the relational structure to be embedded within object positions, it could have been any of the underlying factors. The only requirement is that, for each super-structure, there is a projection that describes a stereotyped manifold in object factor space.

\section{Model architecture}

\subparagraph{Inference network.} We use a graph neural network (GNN) comprising:
\begin{itemize}
    \item an edge MLP with [64, 64] units
    \item a node MLP with [128, 128] units
    \item a globals MLP with [256, 256, \( d_{\rel}\)] units, where \(d_{\rel} \) is the dimensionality of \( \rel \).
\end{itemize}

The GNN is applied for two message-passing iterations, following which the globals are used as the variational posterior parameters \( \muq, \sigq \).

\subparagraph{Generative model.}

The generative model takes a sample \( \rel \sim Q(\rel \mid \abs ) \) and passes it through an MLP with [64, 128, 32] units. The output is then decoded for a fixed number of steps (corresponding to the number of slots, K) by an LSTM with 64 units. Finally, each decoded abstract entity \( \hat{\abs}_i \) is passed through a shared MLP with [128, \( d_{\abs} \)] units, where \( d_{\abs} \) is the dimensionality of each entity. 

\section{Object representation pre-processing}

To ensure disentangled object representation, and to not bias the mask learning towards any of the object features in particular, we perform some preprocessing steps to the MONet representations: we first set the empty slots to default values, and whiten the remaining features (across the whole dataset). Setting empty slots to default values eases the pressure on the network to represent empty slots explicitly. A more general approach would be to learn whether a slot was empty or not, but we leave this to further work.

\section{Relational latents characteristics}

\begin{figure}[!t]
    \centering
    \includegraphics[width=0.5\columnwidth]{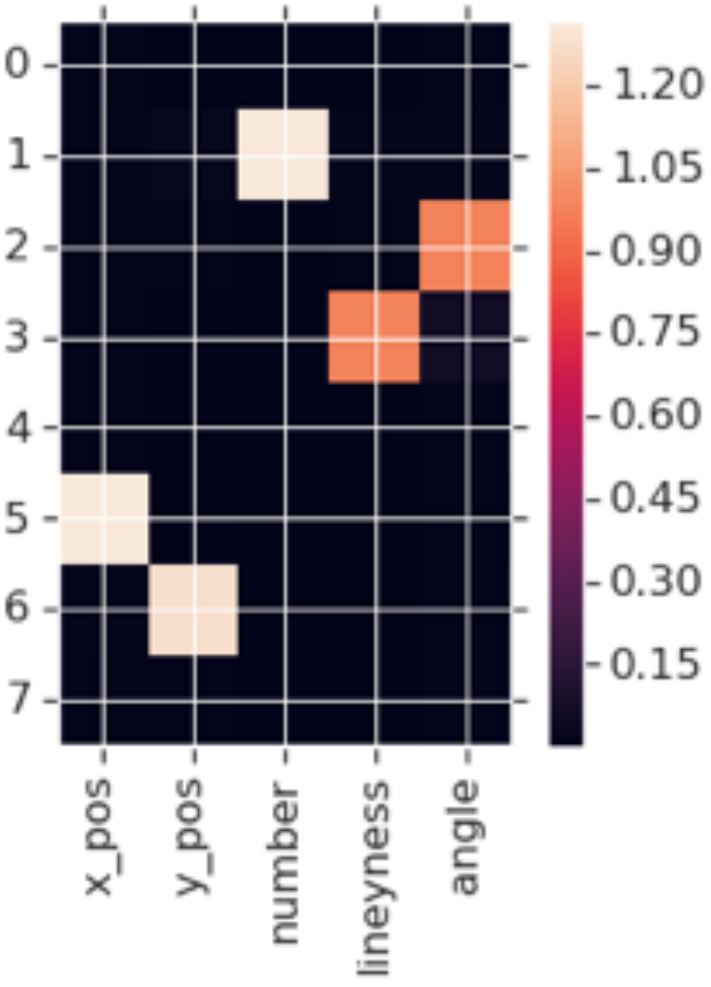}
    \caption{\textbf{\modelname{} learns disentangled relational factors.} Mutual information between \modelname{} latents and ground truth factors.}
    \label{fig:mig}
\end{figure}

To characterise the disentanglement of the relational latents, \( \rels \), we analyse the mutual information between latent dimensions of \( \rels \) and ground truth relational factors (from the dataset generation). These results are shown in Figure \ref{fig:mig}, and demonstrate that \modelname{} learns disentangled relational knowledge.

\section{SCAN}
\begin{figure}[!t]
    \centering
    \includegraphics[width=0.9\columnwidth]{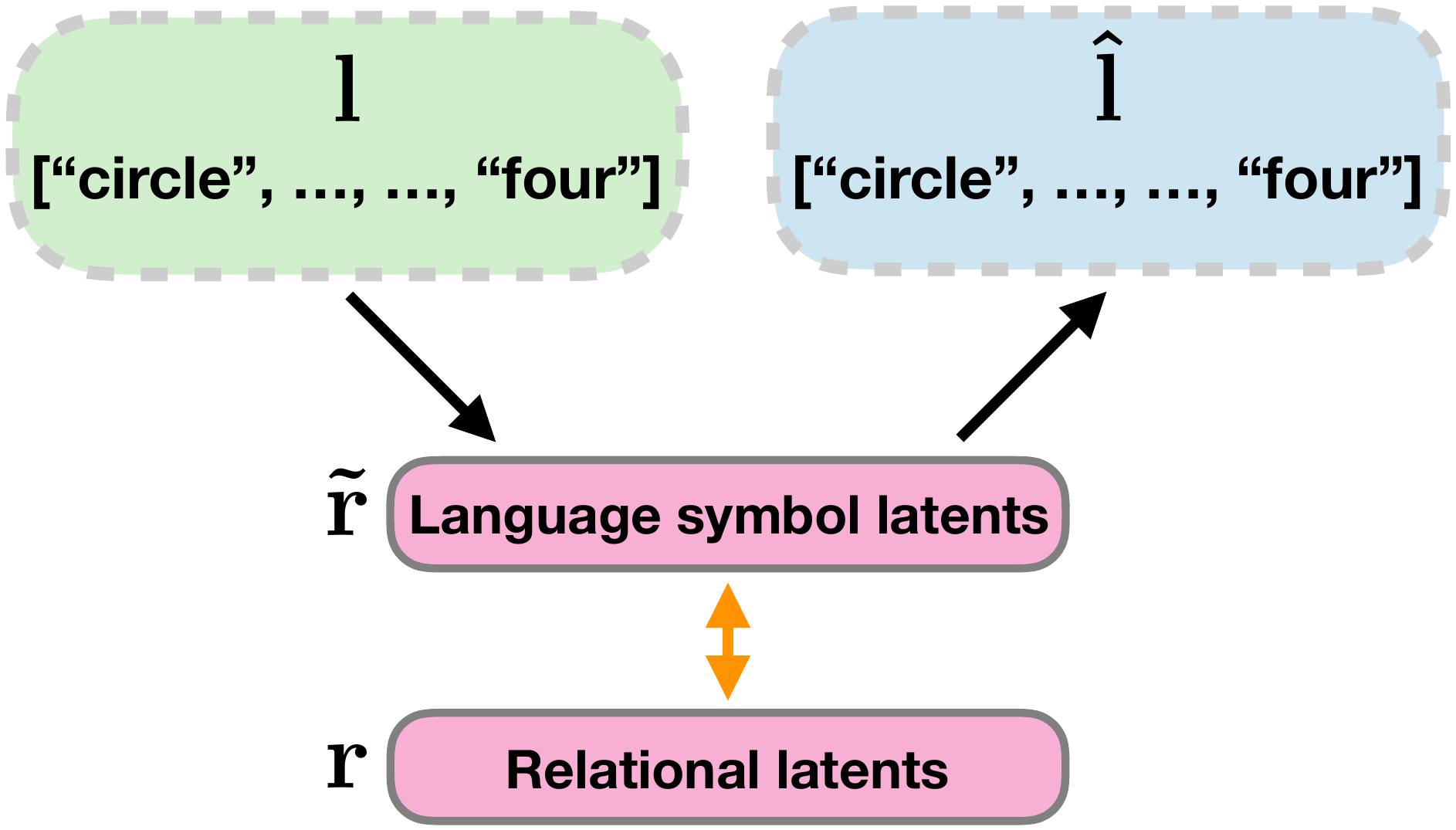}
    \caption{\textbf{SCAN architecture.} Scan learns to encode a language symbol, \( \lan \), to the same latent space as \modelname{}. This forms an effective association between language symbols and relational abstractions.}\label{fig:scan}
\end{figure}

\begin{figure}[!t]
    \centering
    \includegraphics[width=0.8\columnwidth]{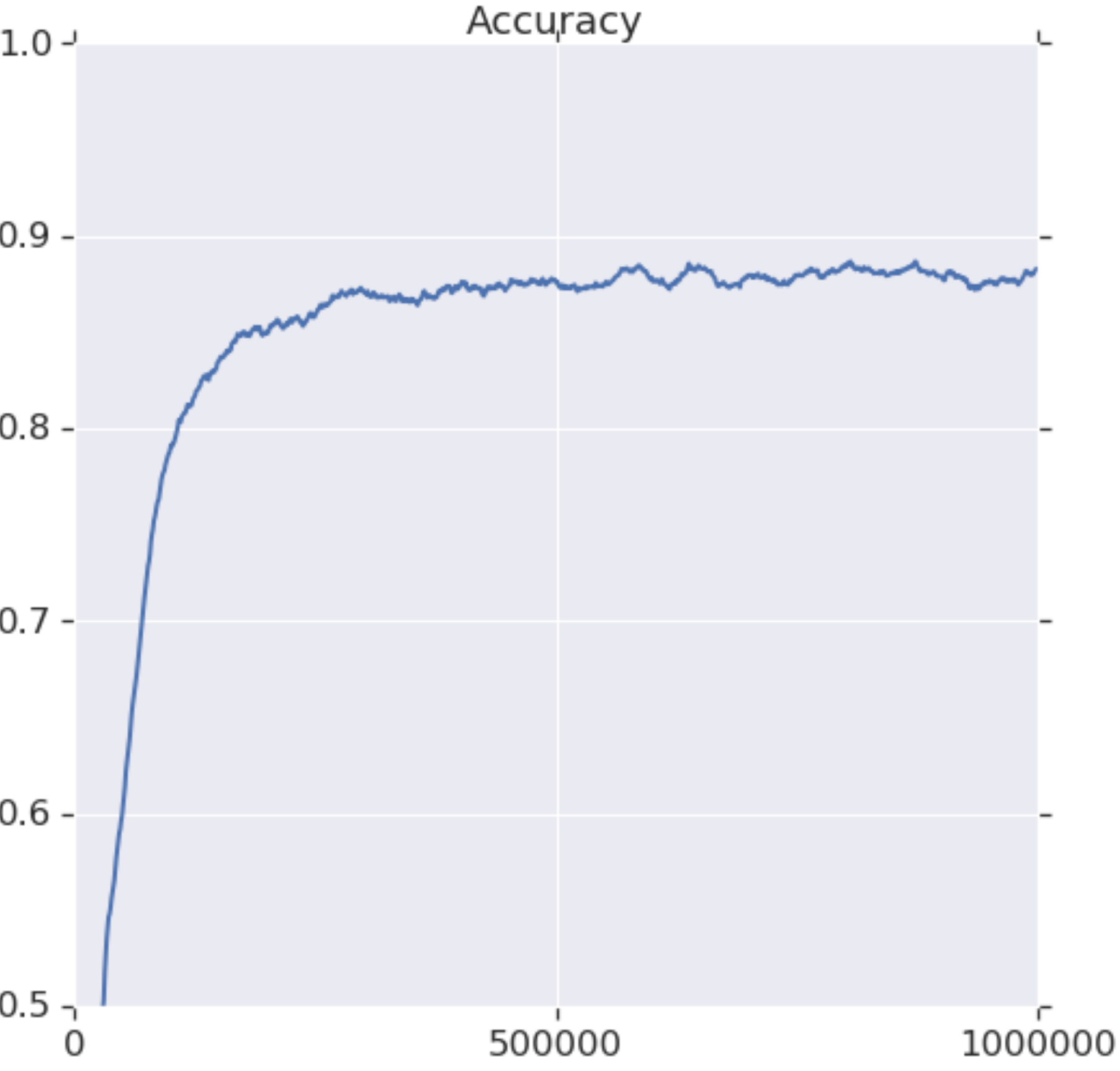}
    \caption{\textbf{SCAN test accuracy} of symbol description from a visual scene's encoded relational factors.}
    \label{fig:scan_acc}
\end{figure}

We now describe a method for associating language symbols to abstractions over our relational representations. We adapt SCAN \cite{Higgins2018}, short for Symbol-Concept Association Network, a network architecture that binds language symbols to abstractions over disentangled image representations (concepts). E.g. the language symbol, \( \lan \), for `straight' will be associated to relational representations, \(\rel\), for any images with objects in a straight line. To facilitate zero-shot composition, we choose to encode the symbols in a multi-hot, disentangled format, such that after binding `straight' and `long' separately, the system can immediately understand what `straight, long' means - a relational representation with latents, \(\rel\), corresponding to any number of objects in a straight line. 

SCAN consists of a VAE taking a language symbol, \( \lan \), as input, and embedding it in a latent space \( \rels \) via the posterior \( Q_{scan} ( \rels \mid l) \). The posterior is a diagonal Gaussian with mean, \( \mus\), and standard deviation, \(\sigs\), parameterised by neural networks. Learning proceeds via a \(\beta\)-VAE loss combined with a reverse-KL term that encourages \( Q_{scan} ( \rels \mid \lan ) \) to match \( Q ( \rel \mid a) \). Since only the curviness latent dimension has a consistent value when paired with `circle' across several input examples, SCAN learns the same value in \( \rels \) and a high precision \(\sigs\) for the corresponds latent dimension, while other latent dimensions will have low precision \(\sigs\). Now if `circle' is provided as an input (instruction), only the latent dimension encoding curviness should have high precision. This is equally true for \textit{compositional} language instructions; if `circle, four' is provided, both the curviness and number dimension should have high precision.
We show an example of the SCAN architecture in the context of our relational latents in Figure~\ref{fig:scan}.

After this weak supervised learning by pairing symbolic instructions with superlatents, \(\rels\), from corresponding examples of scenes, we should be able to categorise, using language symbols, the relational elements of each scene. In Figure~\ref{fig:scan_acc}, we demonstrate good classification accuracy on held out data.




\end{document}